\documentclass{llncs}
\usepackage[hyphens]{url}
\usepackage{hyperref}

\usepackage[fleqn]{amsmath}
\usepackage{graphicx}
\usepackage{xcolor}
\hypersetup{
    colorlinks,
    linkcolor={red!50!black},
    citecolor={blue!50!black},
    urlcolor={black!50!black}
}

\begin{document}

\title{We used Neural Networks to Detect Clickbaits: You won't believe what happened Next!}

\author{Ankesh Anand\inst{1} \and Tanmoy Chakraborty\inst{2} \and Noseong Park\inst{3}}
\authorrunning{Ankesh Anand et al.} 
%
\institute{Indian Institute of Technology, Kharagpur, India\\
\email{anandank@mila.quebec},\\
\and
University of Maryland, College Park, USA \\
\email{tanchak@umiacs.umd.edu}
\and
 University of North Carolina, Charlotte, USA\\
\email{npark2@uncc.edu}
}
\maketitle              
\begin{abstract}
Online content publishers often use catchy headlines for their articles in order to attract users to their websites. These headlines, popularly known as {\em clickbaits}, exploit a user's curiosity gap and lure them to click on links that often disappoint them. Existing methods for automatically detecting clickbaits rely on heavy feature engineering and domain knowledge. Here, we introduce a neural network architecture based on {\em Recurrent Neural Networks} for detecting  clickbaits. Our model relies on distributed word representations learned from a large unannotated corpora, and character embeddings learned via Convolutional Neural Networks. Experimental results on a dataset of news headlines show that our model outperforms existing techniques for clickbait detection with an accuracy of 0.98 with F1-score of  0.98  and ROC-AUC of 0.99.

\end{abstract}
\section{Introduction}
``Clickbait'' is a term used to describe a news headline which will tempt a user to follow by using provocative and catchy content. They purposely withhold the information required to understand what the content of the article is, and often exaggerate the article to create misleading expectations for the reader. Some of the example of clickbaits are:
\begin{itemize}
\item ``The Hot New Phone Everybody Is Talking About"
\item ``You’ll Never Believe Who Tripped and Fell on the Red Carpet"
\end{itemize}

Clickbaits work by exploiting the insatiable appetite of humans to indulge their curiosity. According to the Loewenstein's information gap theory of curiosity \cite{loewenstein1994psychology}, people feel a gap between what they know and what they want to know, and curiosity proceeds in two basic steps -- first, a situation reveals a painful gap in our knowledge (that's the headline), and then we feel an urge to fill this gap and ease that pain (that's the click). Clickbaits clog up the social media news streams with low-quality content and violate general codes of ethics of journalism. Despite a huge amount of backlash and being a threat to journalism \cite{pbs}, their use has been rampant and thus it's important to develop techniques that automatically detect and combat clickbaits.

There is hardly any existing work on clickbait detection except Potthast et al. \cite{potthast2016clickbait} (specific to the Twitter domain) and Chakraborty et al. \cite{chakrabortystop}. The existing methods rely on a rich set of hand-crafted features by utilizing existing NLP toolkits and language specific lexicons. Consequently, it is often challenging
to adapt them to multi-lingual or non-English
settings since they require extensive linguistic knowledge for feature engineering and mature NLP toolkits/lexicons for extracting the features without severe error propagation. Extensive feature engineering is also time consuming and sometimes corpus dependent (for example features related to tweet meta-data are applicable only to Twitter corpora).

In contrast, recent research has shown that deep learning methods can minimize the reliance on feature engineering by automatically extracting meaningful features from raw text \cite{collobert2011natural}. Thus, we propose to use distributed word embeddings (in order to capture lexical and semantic features) and character embeddings (in order to capture orthographic and morphological features) as features to our neural network models. 

 In order to capture contextual information outside individual or fixed sized window of words, we explore several Recurrent neural network (RNN) architectures such as Long Short Term Memory (LSTM) , Gated Recurrent Units (GRU) and standard RNNs. Recurrent Neural Network models have been widely adopted for their ability to model sequential data such as speech \cite{graves2013speech} and text \cite{graves2013speech} well. 
 
Finally, to evaluate the efficacy of our model, we conduct experiments on a dataset consisting of clickbait and non-clickbait headlines. We find that our proposed model achieves significant improvement over the state-of-the-art results in terms of accuracy, F1-score and ROC-AUC score. We plan to open-source the code used to build our model to enable reproducibility and also release the training weights of our model so that other developers can build tools on top of them.

\section{Model}

\begin{figure}[!t]
\centering
\includegraphics[height=5cm, keepaspectratio]{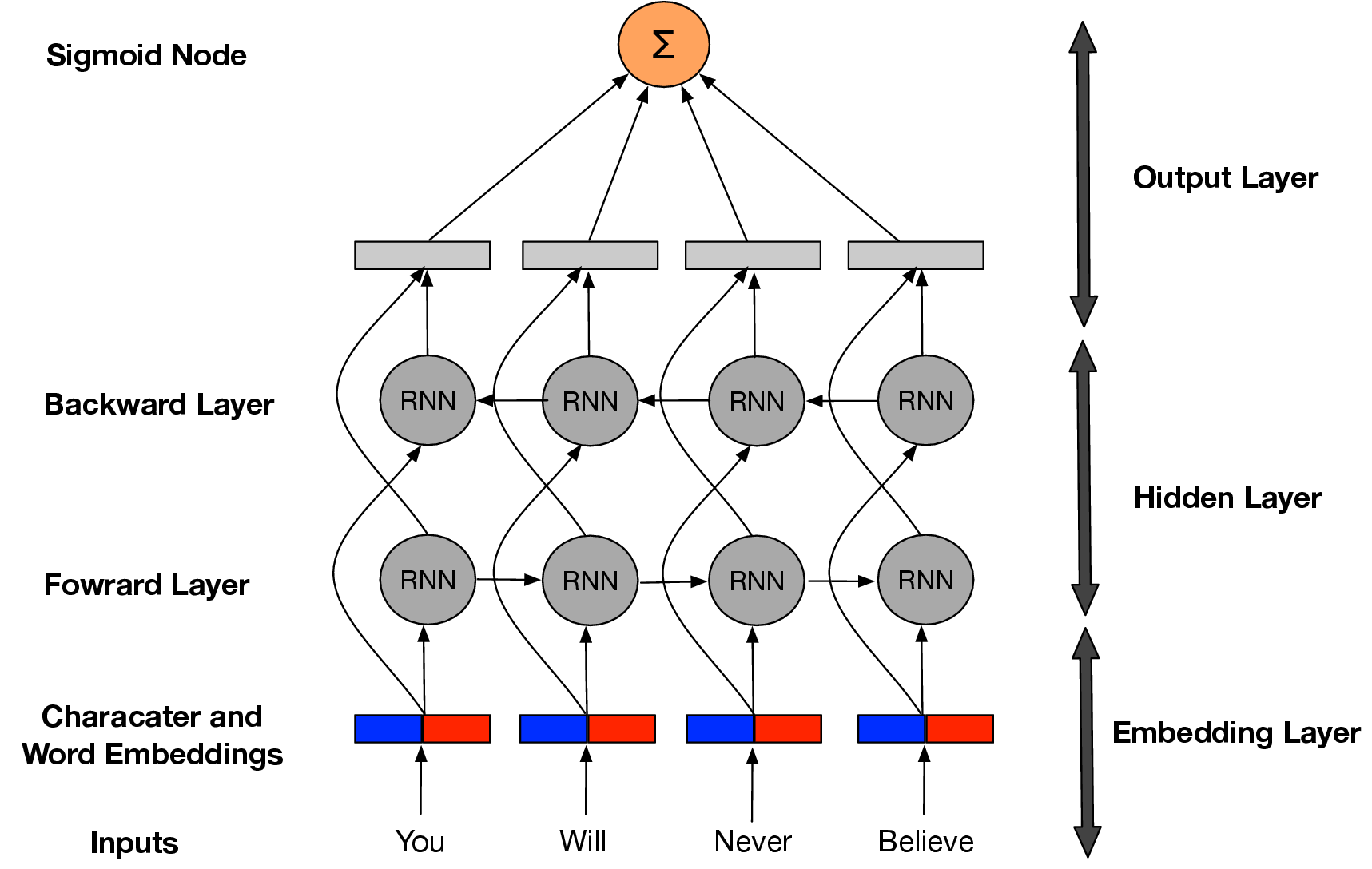}
\caption{BiDirectional RNN architecture for detecting clickbaits}\label{archi}
\vspace{-3mm}
\end{figure}

The network architecture of our model as illustrated in Figure \ref{archi} has the following structure:

\begin{itemize}
\item \textbf{Embedding Layer:} This layer transforms each word into embedded features. The embedded features are a concatenation of the word's Distributed word embeddings and Character level word embeddings. The embedding layer acts as input to the hidden layer.
\item \textbf{Hidden Layer:} The hidden layer consists of a Bi-Directional RNN. We study different types of RNN architectures (described briefly in Section 2.2). The output of the RNN is a fixed sized representation of its input.
\item \textbf{Output Layer:} In the output layer, the representation learned from the RNN is passed through a fully connected neural network
with a sigmoid output node that classifies the sentence as clickbait or non-clickbait.
\end{itemize}

\subsection{Features}
Two types of features are used in this experiment.

\noindent\textbf{Distributed Word Embeddings:} Distributed word embeddings map words in a language to high dimensional real-valued vectors in order to capture hidden semantic and syntactic properties of words. These embeddings are typically learned from large unlabeled text corpora. In our work, we use the pre-trained 300 dimensional word2vec embeddings which were trained on about 100B words from the Google News dataset using the Continuous Bag of Words architecture \cite{mikolov2013efficient}. 

\noindent\textbf{Character Level Word Embeddings:} Character level word embeddings \cite{dos2014learning} have been used in several NLP tasks recently in order to incorporate character level inputs to build word embeddings. Apart from being able to capture orthographic and morphological features of a word, they also mitigate the problem of out-of-vocabulary-words as we can embed any word by its characters through character level embedding. In our work, we first initialize a vector for every character in the corpus. Then we learn the vector representation for any word by applying 3 layers of 1-dimensional CNN \cite{le1990handwritten} with Rectified Linear Unites (ReLU) non-linearity on each vector of character sequence of that word and finally max-pooling across the sequence for each convolutional feature.

\subsection{Recurrent Neural Network Models}
Recurrent Neural Network (RNN) is a class of artificial neural networks which utilizes sequential information and maintains history through its
intermediate layers. A standard RNN has an internal state whose output at each time-step is dependent on that of the previous time-steps. Expressed formally, given an input sequence $x_{t}$, a RNN computes it's internal state $h_{t}$ by:
\begin{eqnarray*}
h_{t} = g(Uh_{t-1} + W_x x_t + b)
\end{eqnarray*}
where $g$ is a non-linear function such as $tanh$. $U$ and $W_{x}$ are model parameters and $b$ is the bias vector.\\

\noindent \textbf{Long Short Term Memory (LSTM):}
Standard RNNs have difficulty preserving long range dependencies due to the vanishing gradient problem \cite{hochreiter1997long}. In our case, this corresponds to interaction between words that are several steps apart. The LSTM is able to alleviate this problem through the use of a gating mechanism. Each LSTM cell computes its internal state through the following iterative process: 
\vspace{-2mm}
\begin{align*}
i_{t} &= \sigma(W_{xi}x_{t} + W_{hi}h_{t-1} + W_{ci}c_{t-1} + b_{i}) \\
f_{t} &= \sigma(W_{xf}x_{t} + W_{hf}h_{t-1} + W_{cf}c_{t-1} + b_{f}) \\
c_{t} &= f_{t} \odot c_{t-1} + i_{t} \odot tanh(W_{xc}x_{t} + W_{hc}h_{t-1} + b_{c}) \\
o_{t} &= \sigma(W_{xo}x_{t} + W_{ho}h_{t-1} + W_{co}c_{t} + b_{o}) \\
h_{t} &= o_{t} \odot tanh(c_{t})
\end{align*}
where $\sigma$ is the sigmoid function, and $i_{t}, f_{t}, o_{t}$ and $c_{t}$ are  the input gate, forget gate, output gate, and memory cell activation vector at time step $t$ respectively. $\odot$ denotes the element-wise vector product. $W$ matrices with different subscripts are parameter matrices and $b$ is the bias vector.\\

\noindent\textbf{Gated Recurrent Unit (GRU):} A gated recurrent unit (GRU) was proposed by Cho et al. \cite{cho2014properties} to make each recurrent unit adaptively capture dependencies of different time scales. Similarly to the LSTM unit, the GRU has gating units that modulate the flow of information inside the unit, however, without having a separate memory cells. A GRU cell computes it's internal state through the following iterative process:
\begin{align*}
z_{t} &= \sigma(W_{z}x_{t} + U_{z}h_{t-1}) \\
r_{t} &= \sigma(W_{r}x_{t} + U_{r}h_{t-1} \\
\tilde{h}_{t} &= tanh(W_{h}x_{t} + U(r_{t} \odot h_{t-1})) \\
h_{t} &= (1-z_{t})\tilde{h}_{t-1}+ z_{t}h_{t}
\end{align*}

where $z_{t}$, $r_{t}$, $\tilde{h}_{t}$ and $h_{t}$ are respectively, the update gate, reset gate, candidate activation, and memory cell activation vector at time step $t$. $W_{h}$, $W_{r}$, $W_{z}$, $U_{r}$ and $U_{z}$ are parameters of the GRU and $\odot$ denotes the element-wise vector product.

In our experiments, we use the Bi-directional variants of these architectures since they are able to capture contextual information in both forward and backward directions.

\section{Evaluation}
\textbf{Dataset:}
We evaluate our method on a dataset of 15,000 news headlines released by Chakraborty et al. \cite{chakrabortystop} which has an even distribution of 7,500 clickbait headlines and 7,500 non-clickbait headlines. The non-clickbait headlines in the dataset were sourced from Wikinews, and clickbait headlines were sourced from BuzzFeed, Upworthy, ViralNova, Scoopwhoop and ViralStories. We perform all our experiments using 10-fold cross validation on this dataset to maintain consistency with the baseline methods.\\

\noindent\textbf{Training setup:}
For training our model, we use the mini-batch gradient descent technique with a batch size of 64, the ADAM optimizer for parameter updates and Binary Cross Entropy Loss as our loss function. To prevent overfitting, we use the dropout technique \cite{srivastava2014dropout} with a rate of $0.3$ for regularization.  During training, the
character embeddings are updated to learn effective representations for this specific task. Our implementation is based on the Keras \cite{chollet2015keras} library using a TensorFlow backend.\\

\noindent\textbf{Comparison of different architectures:}
We first evaluate the performance of different RNN architectures using Character Embeddings (CE), Word Embeddings (WE) and a combination of both (CE+WE). Table 1 shows the result obtained by various RNN models on different metrics (specifically Accuracy, Precision, Recall, F1, and
ROC-AUC scores) after 10-fold cross validation. 
\begin{table}[!t]
\caption{Performance of various RNN architectures after 10-fold cross validation. The 'Bi' prefix means that the architecture is Bi-directional.}
\centering
\scalebox{0.9}{
\begin{tabular}{|l|c|c|c|c|c|}
\hline
\textbf{Model} & \textbf{Accuracy} & \textbf{Precision} & \textbf{Recall} & \textbf{F1-Score} & \textbf{ROC-AUC} \\ \hline
BiRNN (CE) & 0.9629 & 0.9513 & 0.9757 & 0.9633 & 0.9929 \\
BiRNN (WE) & 0.9650 & 0.9722 & 0.9573 & 0.9647 & 0.9935 \\
BiRNN (CE+WE) & 0.9666 & 0.9530 & 0.9787 & 0.9655 & 0.9938 \\ \hline
BiGRU (CE) & 0.9661 & 0.9833 & 0.9482 & 0.9634 & 0.9945 \\
BiGRU (WE) & 0.9769 & 0.9761 & 0.9778 & 0.9770 & 0.9965 \\
BiGRU (CE+WE) & 0.9774 & 0.9662 & \textbf{0.9893} & 0.9776 & 0.9979 \\ \hline
BiLSTM (CE) & 0.9673 & \textbf{0.9849} & 0.9492 & 0.9667 & 0.9950 \\
BiLSTM (WE) & 0.9787 & 0.9759 & 0.9815 & 0.9787 & 0.9970 \\
BiLSTM (CE+WE) & \textbf{0.9819} & 0.9839 & 0.9799 & \textbf{0.9819} & \textbf{0.9980} \\ \hline
\end{tabular}
}
\vspace{-5mm}
\end{table}

We observe that BiLSTM(CE+WE) model slightly outperforms other models, and the BiLSTM architecture in general performs better than BiGRU and BiRNN. If we look at performance of an individual architecture using
three different set of features, model using a combination of word embeddings and character embeddings consistently gives the best results, closely followed by model with only word embeddings.\\  

\noindent\textbf{Comparison with existing baselines:}
Finally, we compare our model with state-of-the-art results on this dataset as reported in Chakraborty et al. \cite{chakrabortystop}. The models reported in \cite{chakrabortystop} use a combination of structural, lexical and lexicon based features. In Table \ref{table_baseline}, we notice  that our BiLSTM(CE+WE) model shows more than 5\% improvement in terms of both accuracy and F1-score and more than 2\% in terms of the ROC-AUC score over the best performing baseline (i.e. Chakraborty et al. \cite{chakrabortystop} (SVM)).

\begin{table}[!h]
\centering
\caption{Comparison of our model with the baseline methods.}\label{table_baseline}
\scalebox{0.9}{
\begin{tabular}{|l|c|c|c|c|c|}
\hline
\multicolumn{1}{|c|}{Model} &
Accuracy & Precision & Recall & F1-score & ROC-AUC \\ \hline
Chakraborty et al. (2016) (SVM) & 0.93 & 0.95 & 0.90 & 0.93 & 0.97 \\
Chakraborty et al. (2016) (Decision Tree) & 0.90 & 0.91 & 0.89 & 0.90 & 0.90 \\
Chakraborty et al. (2016) (Random Forest) & 0.92 & 0.94 & 0.91 & 0.92 & 0.97 \\
\textbf{BiLSTM(CE+WE)} & {\bf 0.98} & {\bf 0.98} & {\bf 0.98} & {\bf 0.98} & {\bf0.99} \\
\hline
\end{tabular}
}

\end{table}
\section{Conclusion}
In this paper, we introduced three different variants of Bidirectional Recurrent Neural Network model for detecting clickbaits using distributed word embeddings and character-level word embeddings. We showed that these models achieve significant improvement over the state-of-the-art in detecting clickbaits without relying on heavy feature engineering. In future, we would like to qualitatively visualize the internal states of our model and incorporate attention mechanism into our model. 

\bibliography{splncs}
\bibliographystyle{splncs}

\end{document}